\documentclass[sigconf,nonacm]{acmart}

\newcommand{\edit}[1]{#1}

\renewcommand\footnotetextcopyrightpermission[1]{}

\AtBeginDocument{%
  }

\copyrightyear{2023}
\acmYear{2023}
\setcopyright{acmcopyright}\acmConference[WSDM '23]{Proceedings of the Sixteenth ACM International Conference on Web Search and Data Mining}{February 27-March 3, 2023}{Singapore, Singapore}
\acmBooktitle{Proceedings of the Sixteenth ACM International Conference on Web Search and Data Mining (WSDM '23), February 27-March 3, 2023, Singapore, Singapore}
\acmPrice{15.00}
\acmDOI{10.1145/3539597.3570425}
\acmISBN{978-1-4503-9407-9/23/02}

\usepackage{multirow}
\usepackage{tabularx, tabulary}
\usepackage{dblfloatfix}
\usepackage{enumitem}
\usepackage{balance}
\usepackage{caption}
\usepackage{subcaption}

\begin{document}

%\title{Inspiration Retrieval from Collections of Visual Assets}
\title{Self-supervised Multi-view Disentanglement for Expansion of Visual Collections}

%% The "author" command and its associated commands are used to define the authors and their affiliations. Of note is the shared affiliation of the first two authors, and the "authornote" and "authornotemark" commands used to denote shared contribution to the research.
\author{Nihal Jain}
\authornote{Work done during an internship at Adobe Research.}
\email{nihalj@cs.cmu.edu}
\affiliation{\institution{Carnegie Mellon University}\city{Pittsburgh}
  \country{United States}}

\author{Praneetha Vaddamanu}
\authornote{Work done when authors were at Adobe Research.}
\email{pvaddama@cs.cmu.edu}
\affiliation{\institution{Carnegie Mellon University}\city{Pittsburgh}
  \country{United States}}

\author{Paridhi Maheshwari}
\authornotemark[2]
\email{paridhi@stanford.edu}
\affiliation{\institution{Stanford University}\city{Stanford}
  \country{United States}}

\author{Vishwa Vinay}
\email{vinay@adobe.com}
\affiliation{\institution{Adobe Research}\city{Bangalore}
  \country{India}}

\author{Kuldeep Kulkarni}
\email{kulkulka@adobe.com}
\affiliation{\institution{Adobe Research}\city{Bangalore}
  \country{India}}

% \author{Nihal Jain\textsuperscript{\rm 1}\footnote{Work done while at Adobe}, Praneetha Vaddamanu\textsuperscript{\rm 1}, Paridhi Maheshwari\textsuperscript{\rm 2}, Vishwa Vinay\textsuperscript{\rm 3}, Kuldeep Kulkarni\textsuperscript{\rm 3}}
% \affiliation{\institution{\textsuperscript{\rm 1}Carnegie Mellon University, \textsuperscript{\rm 2}Stanford University \textsuperscript{\rm 3}Adobe Research}}
% \email{{nihalj, pvaddama}@cs.cmu.edu, paridhi@stanford.edu, {vinay, kulkulka}@adobe.com}

%% By default, the full list of authors will be used in the page headers. Often, this list is too long, and will overlap other information printed in the page headers. This command allows the author to define a more concise list of authors' names for this purpose.
\renewcommand{\shortauthors}{Nihal Jain, Praneetha Vaddamanu, Paridhi Maheshwari, Vishwa Vinay, \& Kuldeep Kulkarni}
%% If needed use a foot or author note to identify equal contribution

\begin{abstract}
Image search engines enable the retrieval of images relevant to a query image. In this work, we consider the setting where a query for similar images is derived from a \textit{collection of images}. For visual search, the similarity measurements may be made along multiple axes, or \textit{views}, such as style and color. We assume access to a set of feature extractors, each of which computes representations for a specific view. Our objective is to design a retrieval algorithm that effectively combines similarities computed over representations from multiple views. To this end, we propose a self-supervised learning method for extracting disentangled view-specific representations for images such that the inter-view overlap is minimized. We show how this allows us to compute the \textit{intent} of a collection as a distribution over views. \edit{We} show how effective retrieval can be performed by prioritizing candidate expansion images that match the intent of a query collection. \edit{Finally, we present a new querying mechanism for image search enabled by composing multiple collections and perform retrieval under this setting using the techniques presented in this paper.\footnote{A version of this paper has been accepted at WSDM 2023.}}
\end{abstract}

\begin{teaserfigure}
\vspace{-0.5em}
\begin{center}
\includegraphics[width=\textwidth]{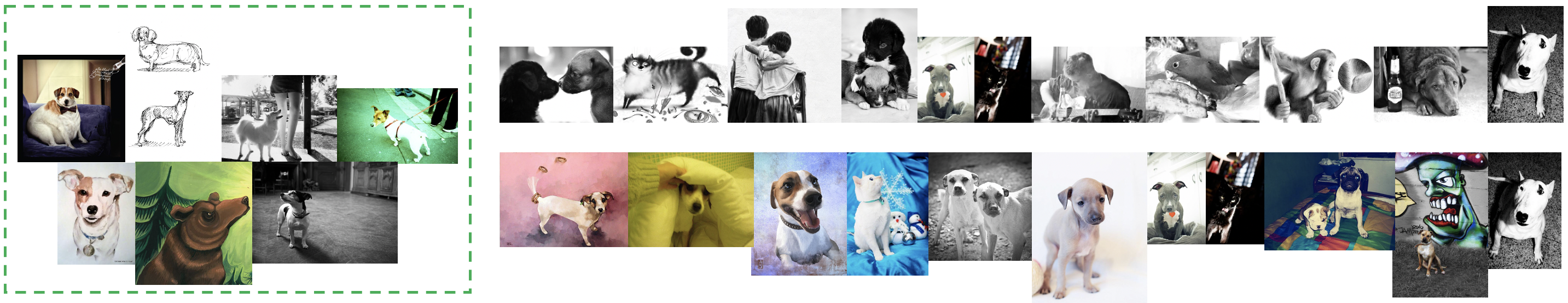}
\end{center}
\vspace{-0.75em}
\caption{\textit{Left}: query collection containing a set of images. \textit{Right}: each row is a ranked list of images that match the query using three notions of image similarity -- objects, style, color composition -- each of which we refer to as a `view'. The top row weighs the views equally. The bottom row (our approach) weighs each view proportional to the inferred intent of the query collection. This enhances relevance (along the primary view - objects) and diversity (along other views - style and color).\\} 
\label{fig:collExpansion}
\end{teaserfigure}

\maketitle

\section{Introduction}
% \njcomment{I have added a list of TODO's based on the WSDM reviewers' comments in WSDM2023-updates.tex. It's a long list, and we may or may not incorporate all feedback.} \pmcomment{I think we should cite the NeurIPS workshop paper as its the same work and will have high overlap on arxiv.} \njcomment{Good point. Though afaik our NeurIPS workshop paper never made it to arxiv. We could cite the original paper anyway.}
The task of image search~\cite{babenko2014neural} requires the definition of specific axes along which image similarities can be computed. The reference standard is the use of embeddings from intermediate layers of convolutional neural networks (CNNs) trained for supervised classification tasks~\cite{krizhevsky2012imagenet}, which have been shown to have superior effectiveness in visual search tasks~\cite{babenko2014neural}. To enable retrieval along specialized notions of image similarity, multiple image feature extractors have been developed. Some examples include shapes within content~\cite{Radenovic_2018_ECCV}, co-occurrences of objects and their relationships~\cite{maheshwari2021scene}, or styles~\cite{ruta2021aladin}. We build on existing image representation methods (e.g. those described above), using the term `view' to refer to a representation capturing one aspect of the content within an image.

We consider the setting of designers working within the context of a visual creation task. As part of ideation, designers typically compile a set of inspirational assets that represents the desired visual characteristics of the target creation -- such a collection is referred to as a ``Moodboard'' \cite{rieuf2015immersive}. In this work, we focus on  \textit{Moodboard Expansion}, i.e., retrieving other visual assets from a corpus of images that match the user's intent as expressed by a moodboard; this is a version of image search where the query is a \textit{collection of images}. Our proposed method for moodboard expansion infers the \textit{intent} of the query collection, and we show how this enables effective retrieval. The intent inference mechanism leverages the fact that, unlike typical retrieval settings where the query is often sparse (e.g. a short textual phrase), in the current setting, we have access to a collection of images. It also provides a convenient visual querying mechanism for our target user personas, who operate in a domain where it might be difficult to express the information need in textual form.

\begin{figure*}[!ht]
\centering
\includegraphics[width=1.0\textwidth]{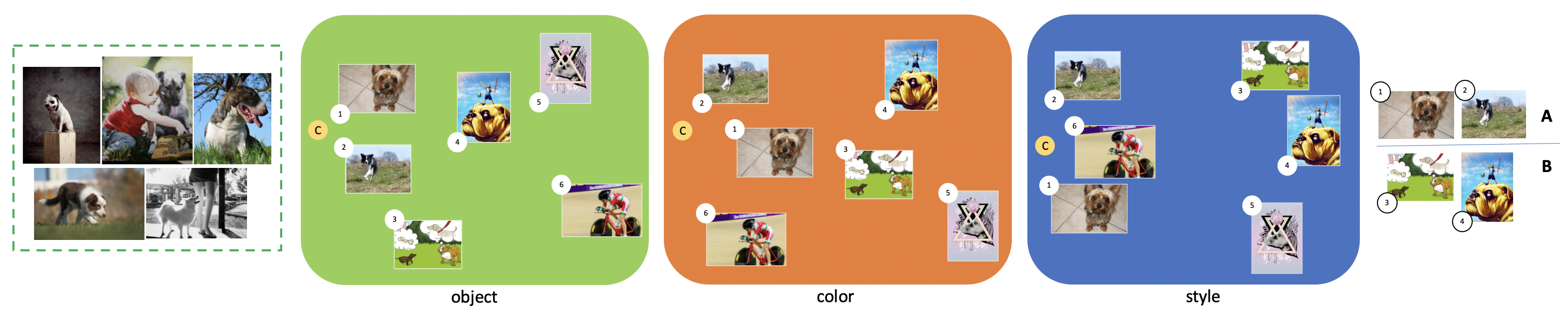}
\caption{A reference example that is intended to motivate the use of multiple views for collection-based retrieval.}
\label{fig:motivation}
\end{figure*}

\edit{A stylized image of our application setting is shown in Figure~\ref{fig:motivation}. On the left is an example moodboard. To surface new candidate additions to the moodboard, we are required to define a notion of similarity for retrieval from a corpus of images -- there could be multiple visual characteristics that we want to consider (e.g., object information, color or style). We formalize each visual characteristic as a separate representation space in which similar images may be found -- we have used the term \textit{views} to refer to these alternative representations. In each subplot of Figure~\ref{fig:motivation}, the point \textit{C} indicates the collection-level representation of the moodboard along the corresponding view. We pictorially depict distances of images from this collection representation in each representation space. Solely using object representations for similarity surfaces the set \textbf{A} on the right, whereas incorporating information along other views may allow the retrieval of content with greater style and color diversity (set \textbf{B}), which may aid more effective visual exploration.}

We describe a self-supervised model, the inputs into which are well known single-view representations, that provides disentangled view-specific representations for individual images. We develop an algorithm that utilizes these disentangled multi-view representations for inferring the intent of a collection and ranks candidate images utilizing the predicted intents. Finally, we provide an empirical study that evaluates our representation learning and image ranking setups. In addition to retrieving images relevant to a query collection, our algorithm ensures diversification of results in the absence of a strong signal along certain views. Figure~\ref{fig:collExpansion} provides an illustrative example of our setup and highlights the desired characteristics of the results.

\edit{Learning view-specific representations for collections of images allows a novel use-case concerning image search: given two (or possibly more) collections of images, we can compose these to hallucinate a new set of images that has selective characteristics from each query collection; we can then retrieve relevant images from an index that match the intent of this hallucinated collection. We present an effective approach to enable this using the techniques presented in this paper. Section~\ref{sec:composing-collections} discusses this in more detail and provides qualitative results achieved using our method.}

Our main contributions are as follows:
\begin{enumerate}[leftmargin=*]
    \item We describe a self-supervised multi-view representation learning method for images. Our model provides a framework to disentangle view-specific information distinct from what is common across views.
    \item We propose an approach to use the view-specific representations output by our model to compute the intent for a collection of images. We validate our intent prediction method via a simulation-based study.
    \item We present experimental results that show how our proposed method leads to more effective visual retrieval than baseline approaches.
    \item \edit{Finally, we propose a novel querying mechanism for image search driven by composing collections of images, and solve this task using other techniques presented in this paper.}
\end{enumerate}

\section{Related Work}

In this section, we review work related to the area of multi-view representation learning of images, as well as retrieval support for visual exploration and ideation activities.

\subsection{Multi-View Representation Learning} \label{sec:multiviewRepresentationLearning}

There has been significant recent interest in representation learning for multimodal content items. Methods for extracting representations from multiple views~\cite{li2018survey,zhao2017multi} usually focus on enforcing alignment across views. Specifically, representations from the different views of the same item are mapped to a shared space, with some notion of closeness enforced between them. These works motivate the need for alignment using a cross-modal task~\cite{tan2019lxmert}, e.g., matching a text caption to the image modality or text-to-image retrieval~\cite{Vo_2019_CVPR}. 

In contrast, our work deals with a naturally multi-view task: even though the views are expected to capture overlapping information, our focus is on extracting the information which is specific to a given view. This notion is closely related to the area of factorized representations. The authors of~\cite{tsai2018learning} motivate the need to separately model factors common across modalities from modality specific factors. We borrow this need, but our definition of \textit{view} is more general since these alternative representations can be derived from the same modality (images in our case). Other authors (e.g.~\cite{tian2020contrastive, daunhawer2021self}) invoke similar intuitions and refer to the desired behaviour as \textit{disentanglement} - we use this word and \textit{orthogonalization} interchangeably in the current paper. 

\subsection{Visual Exploration} \label{sec:visualExploration}

The sub-field of Content-based Image Retrieval (CBIR)~\cite{smeulders2000content} contains many works that are relevant to the topic of the current paper. This includes the need for extraction of the right feature representation for the images, as well as the definition of an appropriate notion of similarity to be used for retrieval. Closest in motivation to our work is ``MindReader''~\cite{ishikawa1998mindreader}, where the authors focus on mechanisms that allow creatives to construct visual queries without resorting to keyword-based interfaces. So as to cover a range of plausible user requirements, MindReader utilizes multiple dimensions (shapes, textures, colors) with the user specifying the relative importance of each. They also account for the correlations between the dimensions, which is also the target of our orthogonalization procedure. 

Our work attempts to tackle moodboard expansion by recommending visual variations that are relevant to the user. Our premise is that the variations that are related to existing assets can encourage ideation. Facilitating design ideation via moodboards has been explored by Rieuf \textit{et al}~\cite{rieuf2015immersive}, though their focus was an immersive interface into the corpus of assets. The recent work of Koch \textit{et al}~\cite{koch2019may} recommends that the exploratory process of ideation be an interactive one, with the system each time refining its view of the user's intent. Solving for the needs of creatives involves a holistic treatment that includes interface, interaction and many more dimensions. The current paper focuses on the quantitative evaluation of a single retrieval iteration. This retrieval is via a weighted similarity across views, where the weights are a prediction of intent of the query collection. The disentangled/orthogonalized multi-view representations of images are central to this process, and are the outputs of our self-supervised model that we describe next. 

\subsection{Compositional Representation  Learning}\label{sec:compositional-learning}
\edit{Recent progress in representation learning has enabled combining representations from multiple simple individual elements to learn representations for complex entities. These simple elements may be from different modalities such as image and text~\cite{vo2019composing, baldrati2022conditioned, anwaar2021contrastive} or the same modality~\cite{redwine2017, nagarajan2018attributes}. Our work focuses on the problem of learning effective representations for collections of images. We demonstrate further utility of these representations by showing how these can be used to achieve composition in a \textit{zero-shot} manner, i.e., without further fine-tuning for this task. Finally, we propose composing collections as a new way of querying images and solve this task using our representations in Section~\ref{sec:composing-collections}.}

\section{Retrieval Using Disentangled Representations} \label{sec:viewSpecificCollectionLevelReperesentations}

\begin{figure}[!h] 
\centering
\includegraphics[width=\columnwidth]{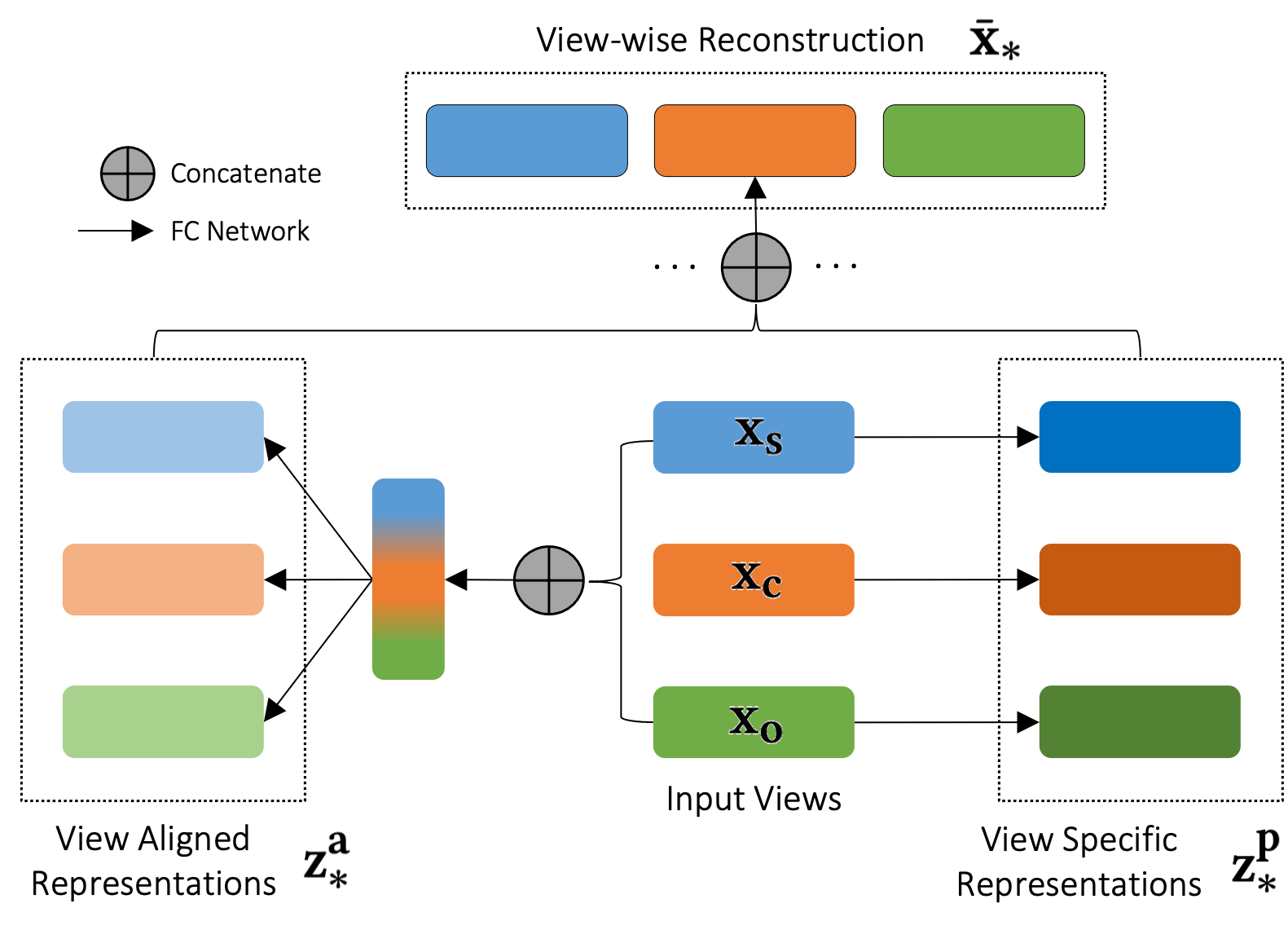}
\caption{Our proposed model. Input features \textbf{$x_*$} are transformed into two outputs: (1) $\mathbf{z^p_*}$ are specific to a view with orthogonalization encouraged amongst them; (2) $\mathbf{z^a_*}$ are aligned to capture common information across views.}
\label{fig:modelDiagram}
\end{figure}

The expansion of a moodboard requires us to understand the motivation behind collating the member images of the collection representing the moodboard. Towards this, we propose the use of representations for collections. We argue (and show empirically, in Section~\ref{sec:expansionOfCollections}) that for effective retrieval, collection level representations need to have independent view-specific components. We next characterize view-specific collection level representations and outline our self-supervised approach to obtain them. 

\subsection{View-Specific Representation Learning} \label{sec:viewSpecificRepresentationLearning}
Our primary premise is that each image can be described by multiple alternative descriptors of its content. For example, images can be projected into a space capturing color semantics such that nearby points have similar color composition; these images can also be projected onto a different space where closeness captures style properties of images. We refer to color and style as views, such that alternative aspects of image content are captured by different views. Our hypothesis is that collection-based retrieval requires effective view-specific understanding of the images comprising the collection. While obtaining shared information across views has been studied in other works (e.g.~\cite{tschannen2019mutual}), in the following sections, we focus on recovering information that is unique to each view. 

\subsubsection{Views and Out-of-the-Box Representations} \label{sec:viewsAndOutoftheboxRepresentations}
The set of views we consider in this paper are specifically chosen based on their central role in visual exploration:
\begin{enumerate}[leftmargin=*]
    \item \textbf{Object}: The object view of images forms the essence of several computer vision tasks such as object detection and segmentation. The ResNet-152 model~\cite{resnet2016} was trained for an object classification task, so, embeddings from its penultimate layer are taken to be the object view; the expectation is that images with similar ResNet embeddings contain visually similar objects.
    \item \textbf{Style}: Our data, described in Section~\ref{sec:datasetDesc}, is derived from an artistic domain, which motivates us to consider the style view. We use the outputs of the ALADIN architecture~\cite{ruta2021aladin}, which was developed to retrieve images based on artistic style similarity.
    \item \textbf{Color}: Since our setting is one of visual discovery, we consider color due to its important role in image retrieval. We utilize the LAB space, with the `$L$' corresponding to luminance and the other two views representing chrominance. As with reference works in this domain~\cite{smith1997visualseek}, the ranges of $L$, $A$ and $B$ values are discretized into bins of width $(10, 10, 10)$, and a color embedding is obtained as a histogram over what fraction of pixels contain a particular LAB value. %This process is known to produce embeddings that are perceptually uniform, i.e., distances in LAB space are proportional to changes in color as perceived by humans. 
\end{enumerate}

We have considered three views for images and associated well-known and state-of-the-art feature extractors --- retrieval using these out-of-the-box representations serve as our baselines. Finally, we note that while we have only described three views, our setup extends naturally to being able to consider a larger enumeration of visual axes. We leave a thorough study of the possible range of visual dimensions and their interactions to future work.

\subsubsection{Disentanglement Model} \label{sec:disentanglementModel}
We intuit that there exist correlations amongst the views: for example, nearest neighbors obtained using ResNet features might also capture similarities in color. Several efforts that study the extraction of features common and specific to views from multi-view data have relied on reconstructing original features from factorized representations~\cite{tian2020contrastive, hazarika2020misa}. Our approach, illustrated in Figure~\ref{fig:modelDiagram}, relies on the same intuition. We factorize the input into two components - the view-aligned representations contain what is common across the views, and the view-specific representations contain information unique to each view. These are then combined to reconstruct the input. %This process reflects our assumption that it should be possible to recover the input from a combination of the view-specific and view-aligned information.

We begin with input representations $\mathbf{x_{m}} \in \mathbb{R}^{d_m}$ for each data point, where view $m \in \mathcal{M}$ has input representations of size $d_m$, and as described previously, we have $\mathcal{M} = \{$object, style, color$\}$ in this paper. Our model processes the input through three neural network pathways:
\begin{enumerate}[leftmargin=*]
    \item \textbf{View-specific}: $\mathbf{z^p_{m}} = \mathcal{F}_m^p\left(\mathbf{x_{m}}\right)$ \\
    \item \textbf{View-aligned}: $\mathbf{z_{m}^a} = \mathcal{F}_m^a\left(\mathbf{u}\right)$, where $\mathbf{u} = \mathcal{F}^u([\mathbf{x_{*}}])$, with $[\mathbf{x_{*}}]$ being the concatenation of all input view representations \\
    \item \textbf{Reconstruction}: $\mathbf{\bar{x}_{m}} = \mathcal{F}_m^r(\mathbf{[\mathbf{z_{m}^p; z_{m}^a}]})$, where $[\mathbf{z_{m}^p; z_{m}^a}]$ is a concatenation of $\mathbf{z_{m}^p}$ and $\mathbf{z_{m}^a}$ \\
\end{enumerate}
Here, all $\mathcal{F}_m^*$ and $\mathcal{F}^u$ are two-layer and one-layer feed-forward networks respectively, where $ReLU$ non-linear activation~\cite{vinod2010rectified} is used between layers.

For ease of notation, we stack $\mathcal{B}$ individual feature vectors into a batch to form data matrices:  $\mathcal{X}_m = [\mathbf{x_{m}}]$ contains input representations, $\mathcal{Z}^p_m = [\mathbf{z^p_{m}}]$ contains view-specific representations,  $\mathcal{Z}^a_m = [\mathbf{z^a_{m}}]$ contains view-aligned representations, and $\mathcal{\bar{X}}_m = [\mathbf{\bar{x}_{m}}]$ contains reconstructed representations for view $m$. All representations that are the output of our models are normalized to be of unit norm, and an inner product between them serves as the definition of similarity between the corresponding embeddings.

We note that $\mathcal{F}_m^a$ is equivalent to multimodal models that project different views to the same underlying space to obtain aligned representations~\cite{baltruvsaitis2018multimodal}. Our intention in the current paper is to extract view-specific representations that capture what is uniquely contained within that view. The role of model components $\mathcal{F}_m^r$ and $\mathcal{F}_m^a$ is to ensure that there is minimal loss of information with respect to the input $\mathbf{x_{m}}$. Therefore, though the focus in the current paper is on $\mathcal{F}_m^p$, the complete architecture illustrated in Figure~\ref{fig:modelDiagram} is required to obtain robust and useful representations.

\subsubsection{Model Fitting} \label{sec:modelFitting}

We define the following loss function that is minimized to estimate the parameters of our proposed model:
\begin{equation}
\label{eqn:loss}
\mathcal{L} =  \lambda_1\cdot{\mathcal{L}_{ali}} + \lambda_2\cdot{\mathcal{L}_{spc}} + \lambda_3\cdot{\mathcal{L}_{inf}} + \lambda_4\cdot{\mathcal{L}_{rec}}
\end{equation}
The $\lambda_i$'s are hyperparameters that control the contribution of the various factors we are attempting to balance. The loss terms in Equation~\ref{eqn:loss} are defined below with respect to batches of size $\mathcal{B}$.
\begin{itemize}[leftmargin=*]

\item \textbf{Inter-view Alignment Loss~} $\left(\mathcal{L}_{ali}\right)$: We use the following objective to align representations from every pair of views $(m, m')$:
    \begin{equation}
        \mathcal{L}_{ali} = \frac{1}{\mathcal{B}} \sum_{(m,m')} \left(\mathcal{B} - trace(\mathcal{Z}^a_m \times \mathcal{Z}^{a^T}_{m'}) \right)
    \end{equation}
This term encourages $\mathcal{Z}^a_m$ and $\mathcal{Z}^a_{m'}$ to be aligned. It is designed to reward an increased similarity between aligned representations of the same data point from different views $m$ \& $m'$ - captured by the diagonal entries of the matrix $\mathcal{Z}^a_m \times \mathcal{Z}^{a^T}_{m'}$.

\item \textbf{Inter-view Orthogonalization Loss~} $\left(\mathcal{L}_{spc}\right)$: This is an orthogonality constraint minimizing the overlap between pairs $(m, m')$:
    % \begin{equation}
    %     \mathcal{L}_{spc} = \sum_{(m,m')} \left\Vert {\mathcal{Z}^{p^T}_m} \times \mathcal{Z}^p_{m'} \right\Vert_2
    % \end{equation}
   
   %%% Is this the correct equation? 
    \begin{equation}
        \mathcal{L}_{spc} = \sum_{(m,m')} \frac{1}{d_m*d_m'} \left\Vert {\mathcal{Z}^{p^T}_m} \times \mathcal{Z}^p_{m'} \right\Vert_F^2
    \end{equation}

%Since $\mathcal{Z}^p_m$ contains unit norm vectors, $\mathcal{L}_{spc}$ is the $L2$-norm of the cross-correlation matrix between pairs of views. 

%%% Correct description?
Since $\mathcal{Z}^p_m$ contains unit norm vectors, $\mathcal{L}_{spc}$ is the squared Frobenius-norm of the cross-correlation matrix between pairs of views.

\item \textbf{Intra-view Information Transfer Loss~} $\left(\mathcal{L}_{inf}\right)$: To prevent degenerate view-specific representations, we introduce a regularization term to retain information content within a view: 
\begin{equation}
    \mathcal{L}_{inf} = \frac{1}{\mathcal{B}} \sum_m \left(\mathcal{B} - trace(\mathcal{X}_m \times \mathcal{Z}^{p^T}_m) \right)
\end{equation}
Since $\mathcal{X}_m$ and $\mathcal{Z}^p_m$ contain unit norm vectors, minimizing $\mathcal{L}_{inf}$ maximizes the cosine similarity between view-specific and input representations for each sample.

\item \textbf{Intra-view Reconstruction Loss~} $\left(\mathcal{L}_{rec}\right)$: The reconstruction loss is the mean squared error between input representations $\mathbf{x_{m}}$ and their estimate $\mathbf{\bar{x}_{m}}$.
    % \begin{equation}
    %     \mathcal{L}_{rec} = \frac{1}{B} \sum_m \left\Vert \mathcal{\bar{X}}_m - \mathcal{X}_m \right\Vert_2^2
    % \end{equation}
    
    %%% Is this the correct equation? 
    \begin{equation}
        \mathcal{L}_{rec} = \sum_m \frac{1}{\mathcal{B}*d_m} \left\Vert \mathcal{\bar{X}}_m - \mathcal{X}_m \right\Vert_F^2
    \end{equation}
    
\end{itemize}

Alternative formulations for these losses are possible, and we leave such explorations to future work.

\subsection{Collection-based Retrieval} \label{sec:collectionbasedRetrieval}
In this section, we describe how the view-specific representations are used for inferring the intent of the collection of images and measuring how true a candidate image is to this intent. 

\subsubsection{Representing a Collection} \label{sec:representingACollection}
We denote a collection of $N$ images as $\mathcal{C}$. Let the view-specific representation for view $m$ of the $i^{th}$ image in $\mathcal{C}$ be denoted as $\mathbf{z_{m,i}^{p}} \forall i \in \{1,\dots, N\}$, which we obtain as outputs of our model described in Section~\ref{sec:viewSpecificRepresentationLearning}. We define the collection-level representation of $\mathcal{C}$ for view $m$ as the mean of the  view-specific representations over its member images:  $\mathbf{C^p_m} = \frac{1}{N}\sum_{i=1}^{N}\mathbf{z^p_{m,i}}$. Computing a query in this manner is similar to how psuedo relevance feedback is used for image retrieval~\cite{zhou2003relevance}. 

\subsubsection{Intent Computation} \label{sec:intentComputation}
Given a collection $\mathcal{C}$, we are interested in inferring why its member images were brought together. We model this by characterizing the intent as being proportional to the degree of homogeneity of images in the collection along that view. We obtain the raw intent of a collection with respect to view $m$ as the average similarity along view $m$ across all pairs of images in the collection:
\begin{equation} \label{eqn:unnormalizedIntent}
    \hat{\beta}_m = \frac{1}{N\times{\left(N - 1\right)}}\sum_{(i,j)} \mathbf{{z}^p_{m,i}} \cdot \mathbf{{z}^p_{m,j}}
\end{equation}
Note that the summand in Equation~\ref{eqn:unnormalizedIntent} computes the average cosine similarity since the output view-specific vectors are normalized. To ensure that these raw intent weights are comparable across views, we standardize them using statistics obtained from each view's embedding space: 
\begin{equation}
    \beta_m = \frac{\hat{\beta}_m - \mu_m}{\sigma_m}
\end{equation}
where $\mu_m$ and $\sigma_m$ are the mean and standard deviation of the pairwise similarities between all pairs of images from the dataset measured along view $m$. Finally, our definition of intent is a normalization across views so that the intent weights sum to $1$: 
\begin{equation} \label{eqn:alpha_m}
\alpha_m = \displaystyle\frac{exp(\beta_m)}{\displaystyle\sum_{m^\prime}exp(\beta_{m^\prime})}
\end{equation}

\subsubsection{Weighted Similarity for Retrieval} \label{sec:weightedSimilarityRetrieval}
Given a collection $\mathcal{C}$, we are interested in ranking a corpus of images ${\mathcal{D}}$ in decreasing order of relevance to $\mathcal{C}$. Given a candidate image $d \in \mathcal{D}$, we obtain its view-specific representations $\mathbf{d^p_m}$ as outputs of our model. We then assign a score to $d$ using a weighted similarity metric as:
\begin{equation}
\label{eqn:score}
    score\left({\mathcal{C},d}\right) = \displaystyle\sum_{m}{\alpha_m \times {sim\left(\mathbf{C^p_m, d^p_m}\right)}}
\end{equation}
where $\mathbf{C^p_m}$ is the the view-specific representations for the collection. And, $\alpha_m$ is computed as in Equation~\ref{eqn:alpha_m} and $sim\left(\mathbf{a, b}\right)$ is a measure of similarity between $\mathbf{a}$ and $\mathbf{b}$. For input style and color representations, we use the inverse of the $L2$ distance between $\mathbf{a}$ and $\mathbf{b}$ as the distance measure~\cite{maheshwari2020learning, ruta2021aladin}. For other representations, we use $sim\left(\mathbf{a}, \mathbf{b}\right) = \mathbf{a} \cdot \mathbf{b}$.  

Equation~\ref{eqn:score} reflects our idea that views corresponding to $\mathcal{C}$'s intent should be given a higher weight while ranking by relevance. Finally, we obtain a ranked list $\mathcal{R}$ by sorting $\mathcal{D}$ in decreasing order of $score({\mathcal{C}, d})$ values $\forall d \in \mathcal{D}$. We discuss the evaluation of $\mathcal{R}$ in Section~\ref{sec:expansionOfCollections}.

\section{Experimental Setup} \label{sec:datasetDesc}
We use the Behance-Artistic-Media dataset (BAM)~\cite{wilber2017bam}, a publicly-available dataset of artistic images. In particular, we use the crowd-annotated subset of BAM, containing $331{,}116$ images (after filtering out broken links). Several images in BAM are annotated with one or more of $3$ \textit{attributes} -- (i) \texttt{content} (associated with $143{,}480$ images), (ii) \texttt{media} ($60{,}225$ images), and (iii) \texttt{emotion} ($24{,}844$ images). Each attribute corresponds to multiple \textit{attribute classes} into which the image may be categorized. Table~\ref{tbl:datasetStats} shows the distribution of images among all attributes and classes in BAM. For each image in BAM, we obtain the outputs of the out-of-the-box feature extractors described earlier -- ResNet (object), ALADIN (style), and LAB Histogram (color) -- as our model inputs. We divide the dataset into train, validation, and test sets, maintaining a $6:3:1$ ratio. While the validation set is used for hyperparameter tuning, images in the test set are set aside to enable the evaluation of our model on the collection expansion task.

\begin{table}[!ht]
\caption{Attributes and Attribute Classes in the BAM dataset.}
\label{tbl:datasetStats}
\centering
\begin{tabular}{@{}ccl@{}}
\toprule
\textbf{Attribute} & \textbf{\# Images} & \multicolumn{1}{c}{\textbf{Attribute Classes}} \\\midrule
\texttt{content} & $143{,}480$ & \begin{tabular}[c]{@{}l@{}}bicycle, bird, building, cars, cat, \\ dog, flower, people, tree\end{tabular} \\\midrule
\texttt{media}   & $60{,}225$  & \begin{tabular}[c]{@{}l@{}}3D graphics, vector art, watercolor,\\pencil sketch, comic, pen ink, oil paint \end{tabular} \\\midrule
\texttt{emotion}   & $24{,}844$    & happy, gloomy, scary, peaceful \\\bottomrule
\end{tabular}
\end{table}

\subsection{Simulating Collections}\label{sec:simulatingCollections}
To evaluate the performance of our approach on the task of moodboard expansion, we simulate the gathering of moodboards of images with known intent using the attribute labels in BAM. By picking a subset of images that are all annotated with the same attribute class, we obtain a collection of images that are similar with respect to that characteristic. For example, gathering a collection of dog images may be simulated by picking a set of images from BAM with the label \texttt{content}~=~ \textit{`dog'}. We refer to such simulated collection  as \texttt{\{attribute\}}-type collections, where \texttt{\{attribute\}}~$\in$~\{\texttt{content, media, emotion}\}. As another example, a sample of images tagged with \texttt{emotion}~=~\textit{`happy'} may be taken to represent an \texttt{emotion}-type collection. 
Given a collection of a known attribute type, we retrieve additional candidate images that are relevant to the collection using the method described in Section~\ref{sec:weightedSimilarityRetrieval}. To judge the relevance of retrieved results, we compute ranking metrics on the top-$100$ retrieved results using the attribute types as labels. 

\subsubsection{Ground Truth Intents} \label{sec:groundTruthIntents}
There are implicit correlations between the attribute labels of BAM and the views we have considered. For each attribute, we identify the view that we expect the attribute to have a high correlation with. We state the following associations between views and types of collections:
\begin{itemize}
    \item \texttt{content}-type collections have high object intent
    \item \texttt{media} and \texttt{emotion}-type collections have high style intent
\end{itemize}

Note that this is knowledge we possess about the dataset, these associations are not made available to the model, which only has access to the input feature extractors for each view. The purpose of intent inference would be to recover the view correlated with the collection's attribute-type as having the highest weight. We demonstrate this via empirical experiments in Section~\ref{sec:evaluatingIntentInferenceFromCollections}. 

\subsection{Metrics} \label{sec:metrics}
Our evaluation setup comprises of two phases: (i) intrinsic: we assess the quality of view disentanglement achieved using our approach, and (ii) extrinsic: we evaluate the effectiveness of our model by computing relevance metrics for the retrieved results. 

\begin{figure*} 
\centering
\includegraphics[width=1.0\textwidth]{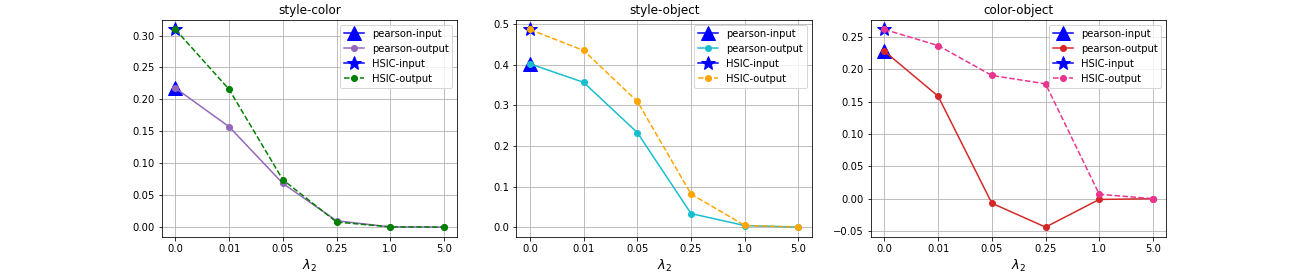}
\caption{Inter view correlations with varying $\lambda_2$ and  fixed \textmd{$\lambda_i$}. As expected, both Pearson and HSIC drop with increasing $\lambda_2$.}
\label{fig:interChannelMI}
\end{figure*}

\subsubsection{\textbf{Evaluating Disentanglement}} \label{sec:evaluatingDisentanglement}
We use two metrics to quantify the disentanglement of view-specific representations:
\begin{enumerate}[leftmargin=*]
\item \textbf{Pearson correlation coefficient}: Let $\mathcal{S} = matsim(\mathcal{P},\mathcal{Q}) \in \mathbb{R}^{n_1 \times n_2}$ represent a matrix of pairwise similarities between entries of $\mathcal{P} \in \mathbb{R}^{n_1 \times d}$ and $\mathcal{Q} \in \mathbb{R}^{n_2 \times d}$ such that $\mathcal{S}(i,j) = sim(\mathcal{P}(i), \mathcal{Q}(j))$. We compute Pearson correlation coefficient between the rows of $matsim(\mathcal{Z}_{m}^p,\mathcal{Z}_{m}^p )$ and $matsim(\mathcal{Z}_{m'}^p,\mathcal{Z}_{m'}^p )$ where $m \neq m'$. This quantifies the inter-view correlation between the pairwise similarities of data points in views $m$ and $m'$. A high Pearson correlation coefficient indicates overlap between the two views, whereas low correlation values indicate that unique aspects are being captured by the two views individually. Similarly, by computing the inter-view correlation between the rows of $matsim(\mathcal{X}_{m},\mathcal{X}_{m})$ and $matsim(\mathcal{X}_{m'}, \mathcal{X}_{m'})$, we obtain measures of overlap between the views when they are represented using input representations. Further, the intra-view correlation between the rows of $matsim(\mathcal{X}_{m},\mathcal{X}_{m})$ and $matsim(\mathcal{Z}_{m}^p,\mathcal{Z}_{m}^p)$ informs us about the deviation of the output view-specific representations from the input representations.
    
\item \textbf{Hilbert-Schmidt Independence
Criterion (HSIC)}: We compute the normalized HSIC metric~\cite{tsai2018learning} as a proxy for the mutual information (MI) between view representations:
\begin{equation} \label{eqn:HSIC}
    HSIC\left(\mathcal{Y}_m, \mathcal{Y}_{m'}\right) = \frac{trace\left({\mathbf{K_{m}HK_{m'}H}}\right)}{\left\Vert{\mathbf{HK_{m}H}}\right\Vert_2\left\Vert\mathbf{{HK_{m'}H}}\right\Vert_2}
\end{equation}
where $\mathbf{K_{m}} = matsim(\mathcal{Y}_m, \mathcal{Y}_m)$, and $\mathbf{H} = \mathbf{I} - {\frac{1}{n}\mathbf{11^T}}$ if $\mathbf{K_{m}} \in \mathbb{R}^{n\times{n}}$. Just as the case with the correlation measure above, we measure inter-view MI between output representations when $\mathcal{Y}_{m} = \mathcal{Z}_{m}^p$ and between input representations when $\mathcal{Y}_m = \mathcal{X}_m$, with $m \neq m'$. We can similarly measure the intra-view mutual information by substituting $\mathcal{Y}_{m} = \mathcal{X}_{m}$ and $\mathcal{Y}_{m'} = \mathcal{Z}_{m}^p$.
\end{enumerate}

\subsubsection{\textbf{Evaluating Expansion of Collections}} \label{evaluatingExpansion}

To quantify the retrieval performance of our approach, we compute the relevance of the \textit{top-k} results of the ranked list $\mathcal{R}$ described in Section~\ref{sec:collectionbasedRetrieval}. We compute the Mean Average Precision (MAP) and Mean Reciprocal Rank (MRR) of $\mathcal{R}$ by using the attribute labels in the BAM dataset (as discussed in Section~\ref{sec:simulatingCollections}) to indicate ground-truth relevance. Specifically, a retrieved image is considered relevant for a query collection if the image belongs to the same attribute class as that used for simulating the query collection.

\section{Experiments and Results}
We validate our method using both intrinsic and extrinsic evaluations using view-specific representations.

\begin{figure}[!h] 
\centering
\includegraphics[width=\columnwidth]{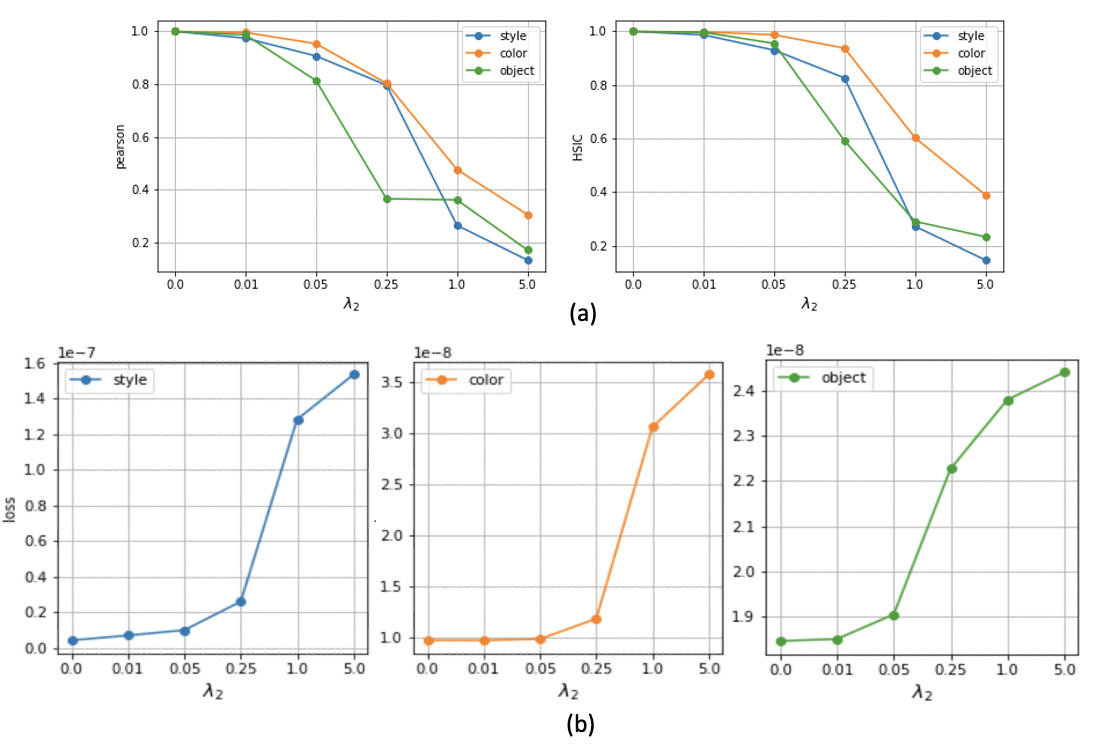}
\caption{(a) Intra-view correlation and MI between input and output representations with increasing $\lambda_2$; (b) Final reconstruction loss for different values of $\lambda_2$. In both cases, varying $\lambda_2$ gives us the control we require, even though the $\lambda_i$ for the other training objective components are kept fixed.}
\label{fig:intraCorrelationLossPlots}
\end{figure}

\subsection{Model Training} \label{sec:modelTraining}
% - Training setup
Our model is trained using the Adam optimizer~\cite{kingma2014adam} with a learning rate of $0.0001$, in batches of size $64$. The other important hyperparameters associated with our model are the $\lambda_i$'s described in Section~\ref{sec:modelFitting}. Since our focus is on learning view-specific representations, which are directly influenced by $\mathcal{L}_{spc}$, we study the effect of varying $\lambda_2$ more closely. We consider $\lambda_2$ from $0.0$ to $5.0$ while keeping the values of the other hyperparameters in the loss function fixed ($\lambda_1=0.001, \lambda_3=0.001$, and $\lambda_4=0.0001$). By sweeping over this operating range for $\lambda_2$, we observe its effect on the disentanglement of input representations. 

Figure~\ref{fig:interChannelMI} visualizes the effect of increasing $\lambda_2$ on the Pearson correlation and HSIC metrics discussed in Section~\ref{sec:evaluatingDisentanglement}. Firstly, when no disentanglement is enforced, i.e., $\lambda_2 = 0$, the metrics computed using input representations and output view-specific representations are almost equal, across all pairs of views. This indicates a complete information transfer between input and output representations. As we increase $\lambda_2$, the inter-view Pearson correlation and HSIC metrics decrease for all pairs of views; this trend is expected for inter-view disentanglement. The decrease in correlation indicates that each output view-specific representation is capturing less information about all other views than their input counterparts.

\begin{figure*}[!h]
\centering
\includegraphics[width=.28\textwidth]{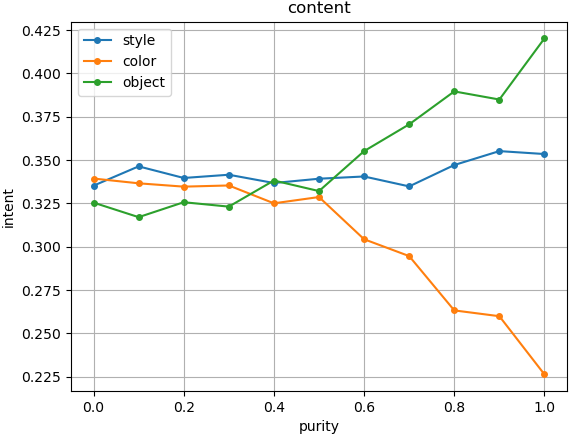}\hfill
\includegraphics[width=.28\textwidth]{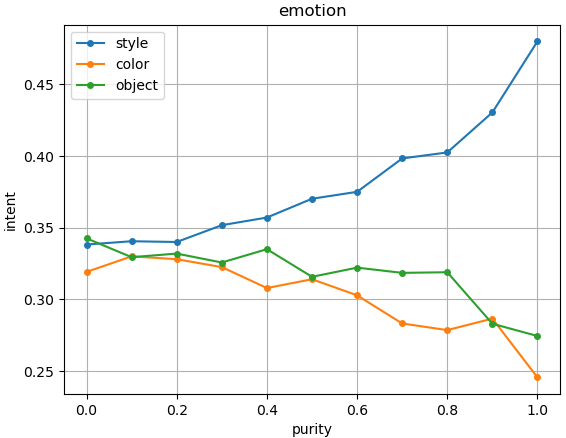}\hfill
\includegraphics[width=.28\textwidth]{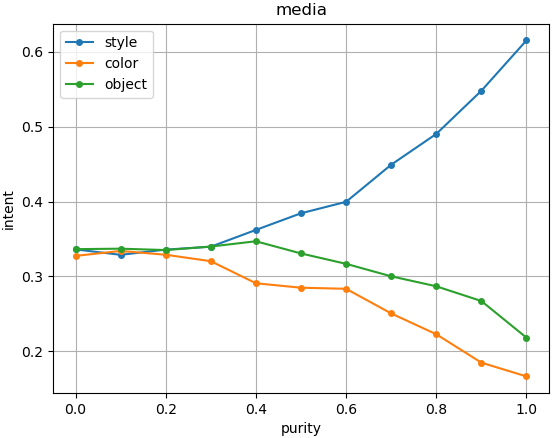}
\caption{View intents as a function of collection purity. Experiments with disentangled representations at $\lambda_2 = 0.05$.}
\label{fig:intentPurity}
\end{figure*}

We are also interested in quantifying the intra-view information retained by our view-specific representations. We measure this as the Pearson correlation and mutual information between the input and output representations for each view. Figure~\ref{fig:intraCorrelationLossPlots}(a) shows the trends as $\lambda_2$ increases. Once again, the decrease in these metrics is expected because the output representations lose information common across views and their overlap with the input decreases. Thus, we note that the intra-view information transfer is influenced by the disentanglement of views because $\lambda_3$, which directly influences the corresponding loss component, was held constant in these experiments. Similarly, the reconstruction loss (Figure~\ref{fig:intraCorrelationLossPlots}(b)) increases despite $\lambda_4$ being held constant in these experiments.

\begin{figure*}[h]
\centering
\includegraphics[width=0.75\linewidth]{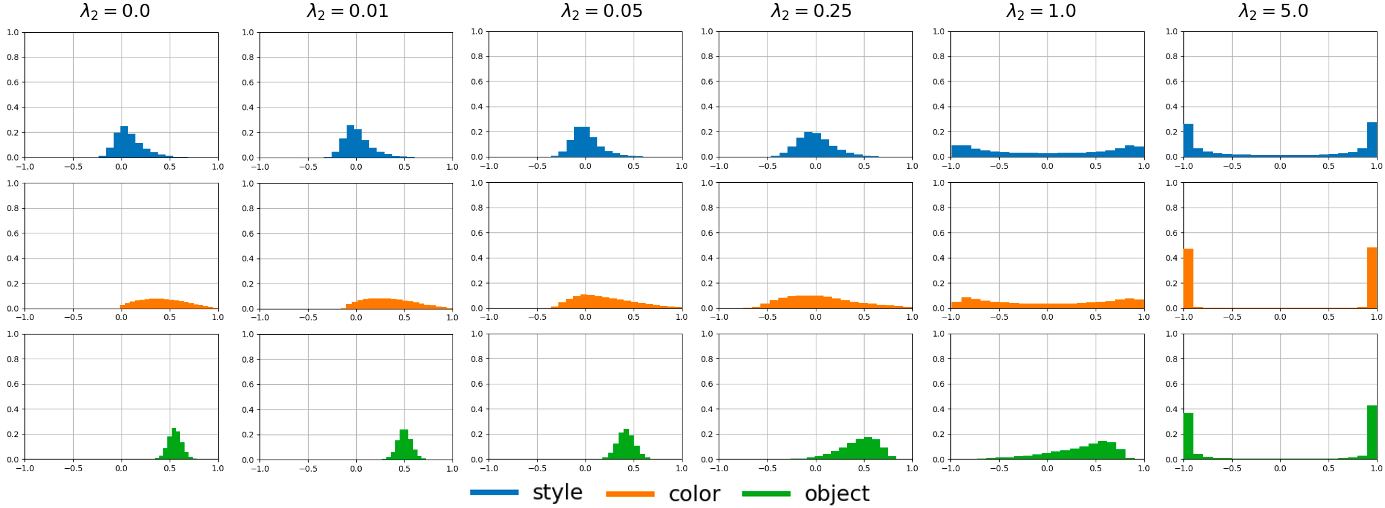}
\caption{Similarities between representations of all pairs of images measured using the output representations of each view. The output representations are obtained by varying $\lambda_2$.}
\label{fig:simHists}
\end{figure*}

We also observe anomalous behavior of the view-specific representations when using large values of $\lambda_2$. For larger values of $\lambda_2$, $\mathcal{L}_{rec}$ converges at relatively higher values indicating difficulty in reconstructing the input representations using these view-specific representations. \edit{In Figure~\ref{fig:simHists}, we show an alternative view of the decreased correlation between input and output representations of the same view, by plotting a histogram of all pairwise similarities between images in the validation set (computed over the disentangled representations) as we sweep over $\lambda_2$. When $\lambda_2 = 0.0$, no disentangling has been enforced and the observed distribution closely resembles the distribution of similarities in the input embedding spaces. With increasing $\lambda_2$ values, we notice a shift in similarity distributions for all the views, indicating departure from the information captured in the input representations. For $\lambda_2 = 5.0$, we observe peaks at similarity values of $-1.0$ and $1.0$, indicating that most representations are either completely orthogonal or nearly identical to others.
This is a degenerate scenario that we would like to avoid. The optimal value of $\lambda_2$ would therefore be somewhere in the middle.} To operate in higher disentanglement regimes, future work may incorporate recently proposed regularization methods (e.g.~\cite{bardes2021vicreg}) to prevent degenerate situations.

We would like to choose hyperparameters based on this intrinsic evaluation, and evaluate the model and learnt representations in the downstream collection-based retrieval task. Since only the relative values of the loss function components matter, we retain $\lambda_1$, $\lambda_3$ and $\lambda_4$ as before and choose $\lambda_2$ based on the trends described above. Choosing a very low value of $\lambda_2$ would not allow us to investigate the benefits of disentanglement, while we have seen that larger values of $\lambda_2$ lead to degenerate behavior. From the HSIC and Pearson correlation values of the inter/intra-view representations, we pick an intermediate value, $\lambda_2 = 0.05$, to evaluate under the collection expansion task. We have also conducted a complete grid search of the hyper-parameters; since they do not add many more insights to the current findings, they have not been reported here.

\begin{table*}[!ht]
\centering
\caption{Evaluation of different representations on the collection expansion task. The first three representations correspond to the baselines of using single view (object, style and color respectively) representations for the images.}
\label{tbl:RankingResults}
\begin{tabular}{ccccccccc}
\toprule
\multirow{2}{*}{\textbf{Representation}} & \multicolumn{3}{c}{\textbf{Attribute-wise MAP}} & \textbf{Aggregate} & \multicolumn{3}{c}{\textbf{Attribute-wise MRR}} & \textbf{Aggregate} \\
\cmidrule{2-4}
\cmidrule{6-8}
& \texttt{content} & \texttt{media} & \texttt{emotion} & \textbf{MAP} & \texttt{content} & \texttt{media} & \texttt{emotion} & \textbf{MRR} \\
\midrule
ResNet          & 0.823     & 0.400   & 0.305    & 0.523 & 0.922   & 0.557   & 0.409   & 0.668 \\
ALADIN          & 0.565     & 0.697   & 0.483    & 0.585 & 0.736   & 0.880	 & 0.708   & 0.800 \\
LAB Histogram   & 0.202     & 0.149   & 0.108    & 0.158 & 0.371   & 0.237   & 0.213   & 0.269 \\
input-uniform   & 0.685     & 0.617   & 0.440    & 0.581 & 0.913   & 0.845	 & 0.660   & 0.809 \\
input-output    & 0.813     & 0.692   & 0.497    & 0.685 & 0.950   & 0.899   & 0.715   & 0.889 \\
output-output   & \textbf{0.857} & \textbf{0.719} & \textbf{0.513} & \textbf{0.713} & \textbf{0.983} & \textbf{0.924} & \textbf{0.797} & \textbf{0.896} \\
\bottomrule
\end{tabular}
\end{table*}

\subsection{Expansion of Collections} \label{sec:expansionOfCollections}
In this section, we evaluate the performance of the output view-specific representations on the task of collection-based retrieval.

\subsubsection{Evaluating Intent Inference} \label{sec:evaluatingIntentInferenceFromCollections}
The first step in expanding a moodboard involves predicting the intent of the collection of images, and we evaluate this component in isolation. We define a \textit{pure} collection where all member images belong to a single attribute class (e.g. \texttt{emotion}~=~\textit{`scary'}). By injecting images belonging to other attribute classes into \textit{pure} collections, we obtain \textit{impure} collections. The fraction of images in a collection that belong to the attribute class used to simulate a collection provides a measure of its \textit{purity}. 

We vary the purity of collections and show that the computed intent weights respond as expected in Figure~\ref{fig:intentPurity}. Each subplot is obtained by simulating collections that contain images belonging to a specified attribute class (indicated as the subplot title) and a given purity level ($x-axis$). For a given collection, the view-specific intents are computed as defined in Section~\ref{sec:intentComputation}. For each purity value, we have plotted the average view-specific intent weights computed over $100$ collection simulations. 

Specifically, when $purity=0$ (the leftmost points), a collection contains a uniform mixture of images from all attribute classes; the intent weights reflect this by being $1/3$ across the three views. As the purities of \texttt{media}-type and \texttt{emotion}-type collections are increased, the style intents increase while color and object intents reduce. Similarly, as we increase the purity of \texttt{content}-type collections, the object intent consistently increases and reaches a maximum value for pure collections. These findings agree with the known attribute-view correlations discussed in Section~\ref{sec:simulatingCollections}, and therefore validate our method for inferring the intent. The disentanglement of view-specific representations is critical to having this behavior --- a rising trend of intent with increasing purity is observed only for the view correlated with a collection's attribute class. 

\subsubsection{Retrieving Images for Collection Queries} \label{sec:relevanceTrends}
Our method for collection-based retrieval involves computation of intent weights and the subsequent retrieval of images using a weighted similarity. It is not necessary that the representations used for computing intent weights (in Equation~\ref{eqn:alpha_m}) be the same as the ones used for computing $sim$ scores for collection-based retrieval (in Equation~\ref{eqn:score}). We therefore consider the following variations in our experiments:
\begin{enumerate}[leftmargin=*]
    \item \texttt{input-uniform} --- Input representations are used for $sim$ score computation and intent weights are uniform across views. This is a multi-view setting without intent inference.
    \item \texttt{input-output} --- Input representations are used for $sim$ score computation while view-specific (output) representations are used for intent weight inference.
    \item \texttt{output-output} --- View-specific (output) representations are used for both computing intent weights and $sim$ scores.
\end{enumerate}

\begin{figure*}[!ht]
\centering

\begin{subfigure}{\textwidth}
\includegraphics[width=.3\textwidth]{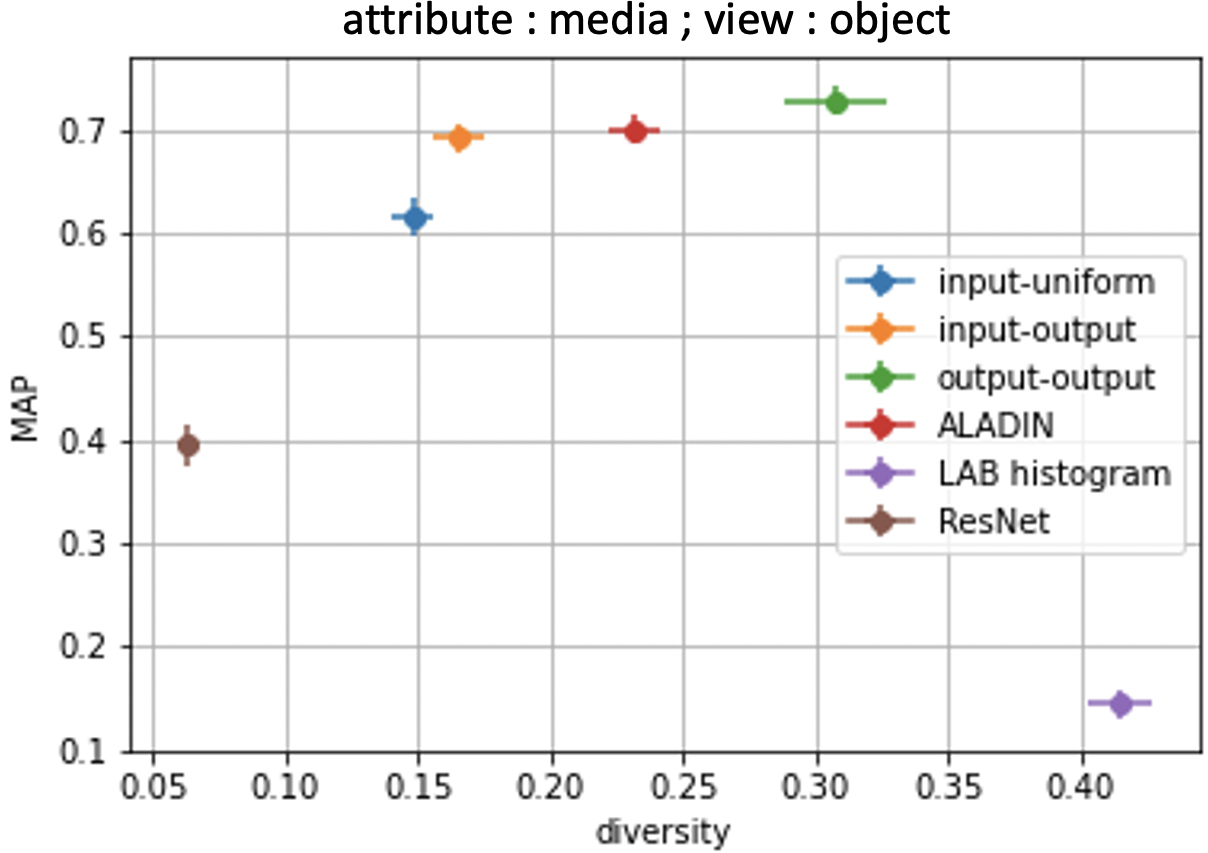}\hfill
\includegraphics[width=.3\textwidth]{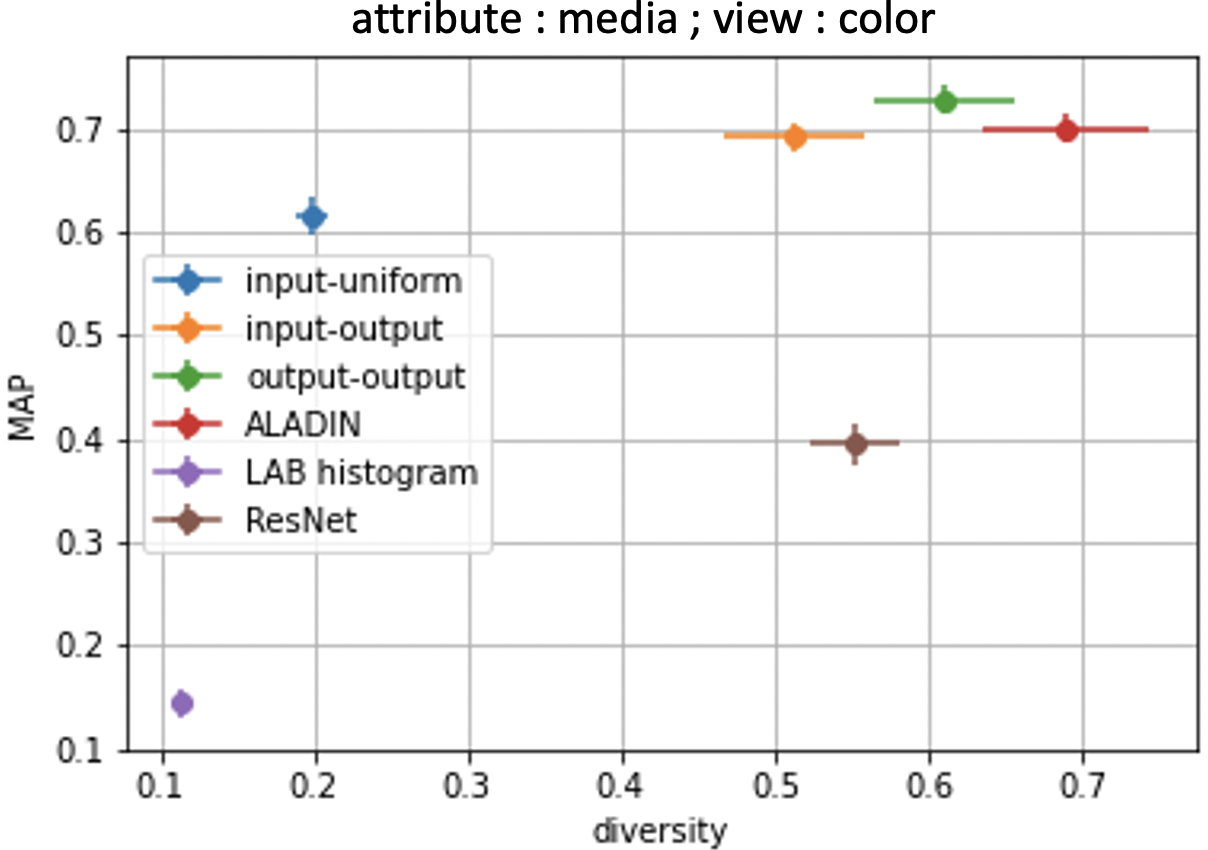}\hfill
\includegraphics[width=.3\textwidth]{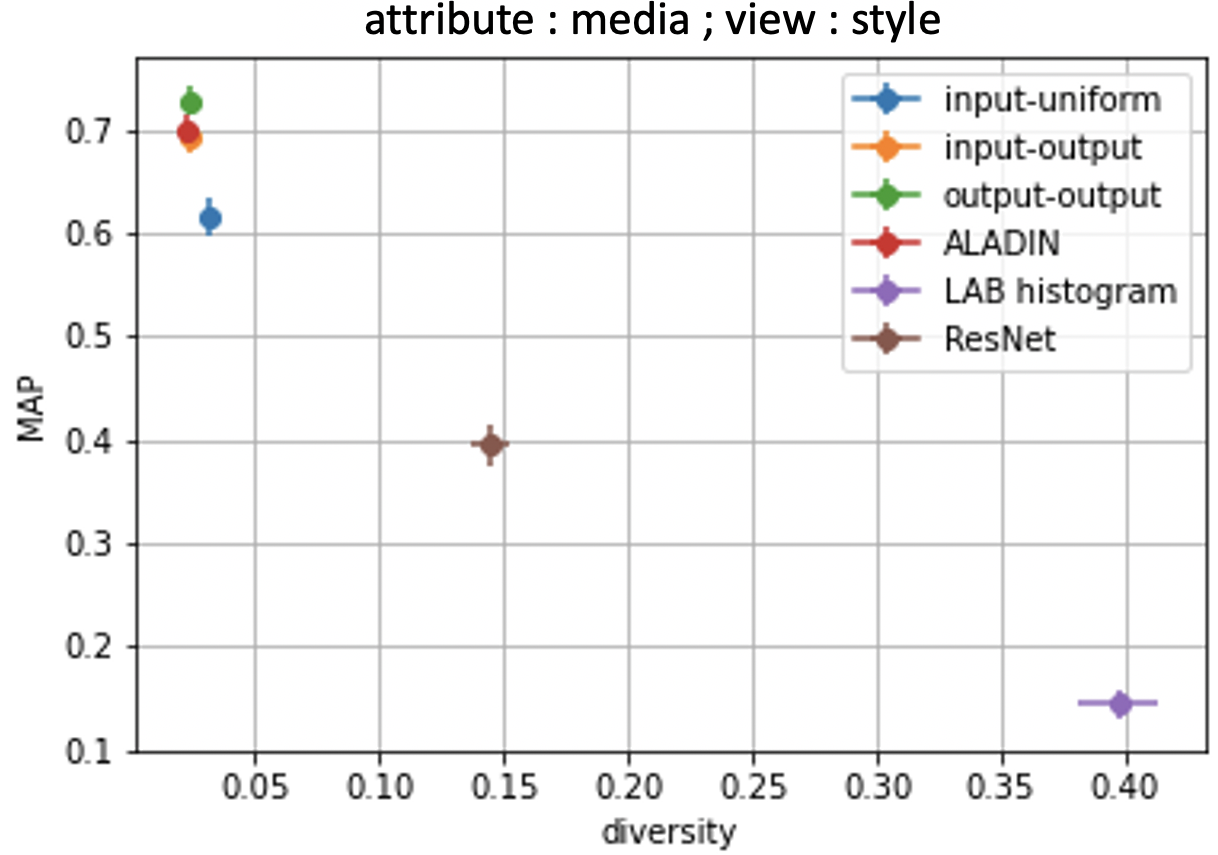}
\caption{MAP-diversity trade-off for \textmd{\texttt{attribute = media}}. $\lambda_2 = 0.05$. Values are means computed over $100$ simulated collections, with the standard error also plotted for both dimensions. Values towards the top-right are desirable, as they indicate that the corresponding configuration provides the dual benefits of increased relevance (MAP) and more effective exploration (diversity). Our full model (green dot) achieves the best performance overall.}
\label{fig:MAPDiversitya}
\end{subfigure}

\begin{subfigure}{\textwidth}
\includegraphics[width=.3\textwidth]{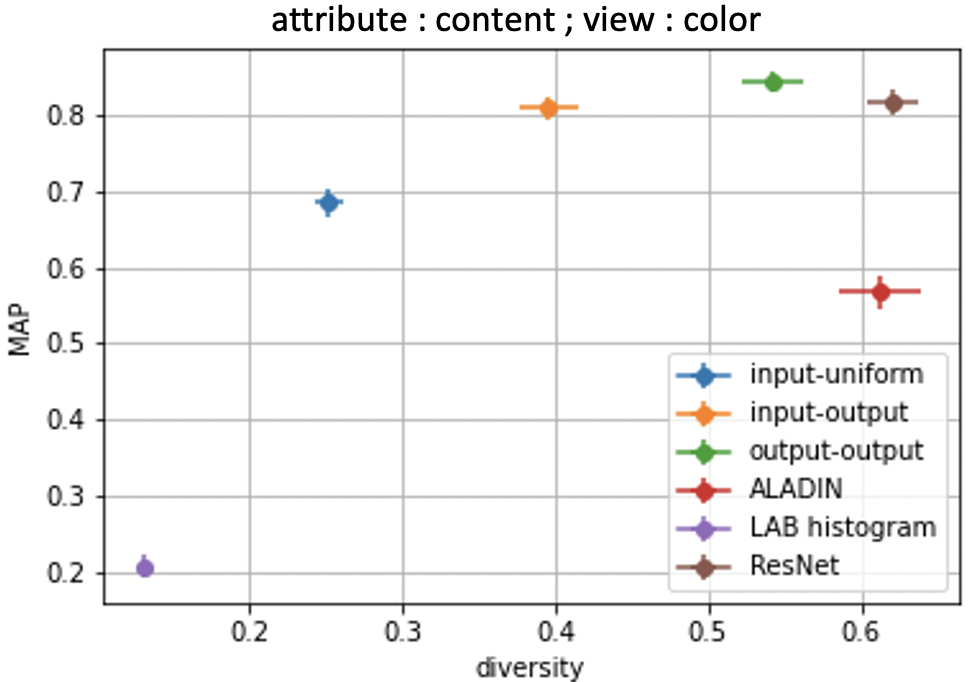}\hfill
\includegraphics[width=.3\textwidth]{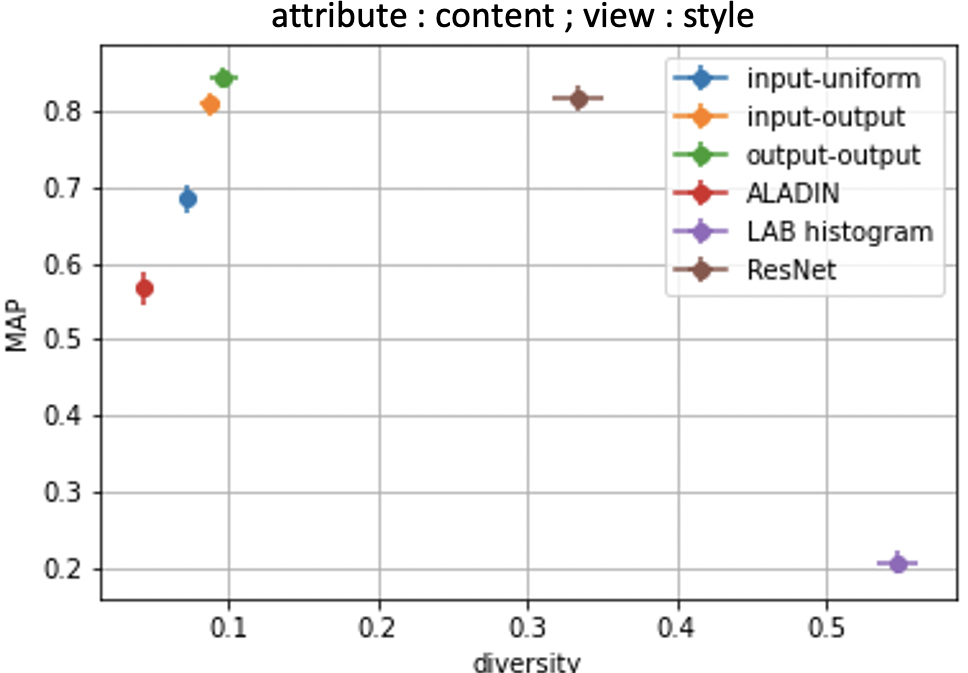}\hfill
\includegraphics[width=.3\textwidth]{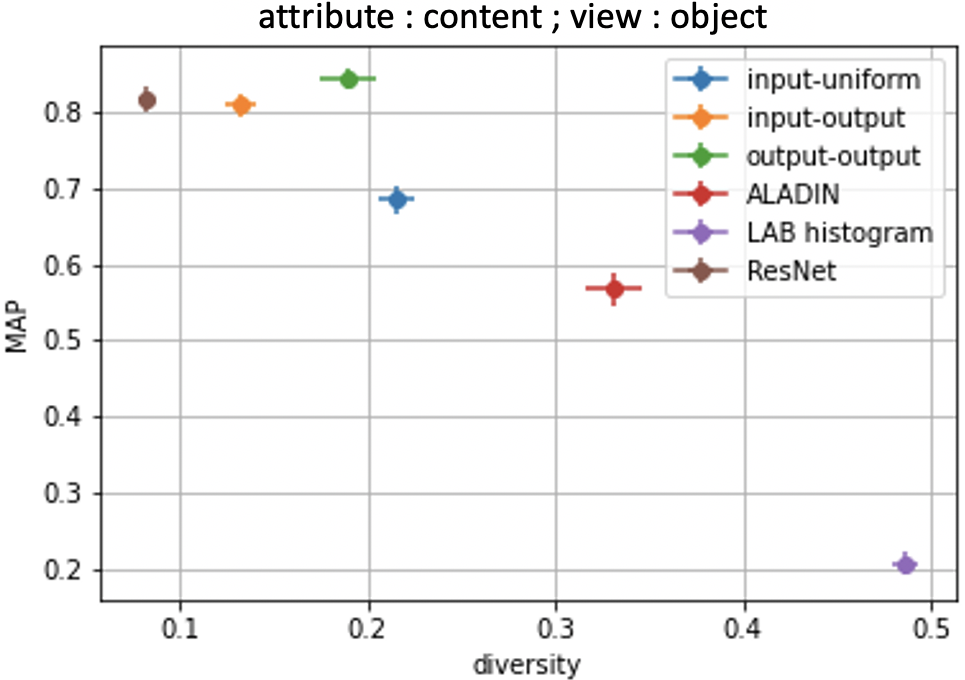}
\caption{The equivalent plot for the \texttt{object} view, i.e., for \texttt{attribute = content}. $\lambda_2 = 0.05$ as before. The \textit{output-output} combination of using representations derived from our our model for collection expansion does the best overall. Using the ResNet representations (expected to be correlated with the object view) provides the most competitive single view MAP.}
\label{fig:MAPDiversityb}
\end{subfigure}
\vspace{-1em}
\caption{\edit{MAP-diversity trade-off for two attribute types.}}\label{fig:MAPDiversity}
\end{figure*}

Table~\ref{tbl:RankingResults} presents MAP and MRR values obtained across our experiments. The first three rows are our baselines: for each view (object, style, and color), we use the corresponding out-of-the-box view representation (ResNet, ALADIN, and LAB Histogram) only for collection expansion via a simple nearest neighbor search without intent inference. The remaining three rows show the results for the variations discussed above. The Attribute-wise MAP \& MRR columns are computed by fixing the attribute while simulating collection queries. Finally, Aggregate MAP \& MRR values are computed by selecting the attribute label for each simulated collection at random and averaging across them. The results reported are the averages over $100$ simulated collections for each configuration. In each case, the number of images in the query collection varies uniformly between $10$ and $30$. We make the following observations about the results shown:

\begin{enumerate}[leftmargin=*]
    \item Among the baselines, for each attribute, the correlated view representation scores the highest Attribute-wise MAP \& MRR. This is especially noticeable in the performance of the ResNet representation for \texttt{content}-type collections, which we expect to have high object intent.
    \item For \texttt{input-uniform}, the Aggregate MAP value is higher than two of three baselines, with Aggregate MRR higher than all three baselines; this indicates that each view provides incremental value when used with others, validating the use of multi-view representations for image retrieval.
    \item \texttt{input-output} outperforms the preceding methods that do not use intent inference; this validates our intent prediction mechanism which is able to selectively invoke the relevant view based on the query collection. 
    \item Finally, disentangled multi-view representations combined with the inferred intent, (\texttt{output-output}), provides the best MAP and MRR in all cases, showing the utility of cross-view disentangled representations.
\end{enumerate}

%\subsubsection{Qualitative Example} \label{qualitativeEvaluation}

%Figure~\ref{fig:collExpansion} shows the retrieved results given a \texttt{content}-type collection with images from the \textit{`dog'} class as query. The top row results were obtained using the \texttt{input-uniform} variant, and the bottom row using the \texttt{output-output} variant. Intent weights computed for the query using view-specific (output) representations were $0.654, 0.209, 0.137$ for the object, style and color views respectively. We make two main observations: (1) when using the disentangled representations for intent computation (\texttt{output-output}), the values of $\alpha_m$ across views is less flat; (2) diversities along the non-intent views are increased even while relevance along the intent view is maintained. That is, the retrieval focuses on returning images of dogs, but ensures that they span a range of styles and colors. In particular, our analysis shows how our intent inference mechanism allows us to control the degree of diversification –- matching the existing images on views of low heterogeneity, but diverging from the query along the other views.

\subsection{Relevance - Diversity Trends}\label{sec:relevance-diversity-trends}
In Table~\ref{tbl:RankingResults}, we show that the ranking effectiveness by using multiple views is on average better than what can be achieved from a single view. When the goal is to provide interesting additions to a designer's moodboard, a core requirement of the retrieval system is to ensure that the user has visibility into the full range of possibilities. This need for exploration is well-studied in the information retrieval community under the notion of diversity~\cite{chen2006less, clarke2008novelty,rodrygo2015search}. 

We are interested in measuring the diversity of results in the returned list of candidate images $\mathcal{R}$ (Section~\ref{sec:weightedSimilarityRetrieval}). Since we are operating in a multi-view space, we can make diversity measurements along each view. Specifically, we measure diversity along view $m$ as $\delta_m = 1/\beta_{m}$, where $\beta_{m}$ is computed as in Section~\ref{sec:intentComputation} by treating $\mathcal{C} = \mathcal{R}$~\cite{ma2010diversifying}. This definition reflects our assumption that intent and diversity are inverse notions of each other. The results from our experiments are shown graphically in Figure~\ref{fig:MAPDiversity}, where each subplot shows the diversity with respect to the specified view on the $x-axis$, and MAP on the $y-axis$. We provide results for \texttt{media}-type collections and \texttt{content}-type collections. 

We make the following observations concerning Figure~\ref{fig:MAPDiversitya}:
\begin{enumerate}[leftmargin=*]
    \item ALADIN shows higher relevance scores than LAB Histogram or ResNet. As \texttt{media}-type collections are anticipated to be correlated with the style view, this behavior is expected. Further, ALADIN has greater diversities along object and color views, and the least diversity along the style view.
    \item Among the variations that use intent weights while ranking, \texttt{input-uniform} has the least diversity along object and color views. This undesired behavior is also expected -- by giving uniform intent, we weigh uncorrelated views more than necessary.
    \item In a multi-view setting, using the output representations solely for intent computation (\texttt{input-output}) leads to higher relevance scores when compared to uniform intent weights. We observe higher diversities along uncorrelated views as desired.
    \item As shown in Table~\ref{tbl:RankingResults}, when using the disentangled view-specific representations for both similarity measurement and intent computation (\texttt{output-output}), we observe the highest relevance scores. In Figure~\ref{fig:MAPDiversitya}, we additionally observe that this scenario produces the highest diversity along the object view, with comparable diversities to ALADIN along the other views.
\end{enumerate}
\edit{Similar observations can be made from Figure~\ref{fig:MAPDiversityb} with respect to the object view as well. A minor difference is that ResNet obtains the highest diversities for uncorrelated views, indicating the alignment between ResNet representations and the object view.}

Thus we show that our weighted nearest neighbor computation enables us to retrieve images that are similar along the view corresponding to the user's intent, while allowing diversity along the other views. Often, the relationship between relevance and diversity is described as a trade-off. The use of our model outputs for computing both intents and similarities leads to MAP values comparable to those observed with the correlated view but with increased diversity along uncorrelated views. 

\begin{figure*}[h]
\includegraphics[width=0.75\textwidth]{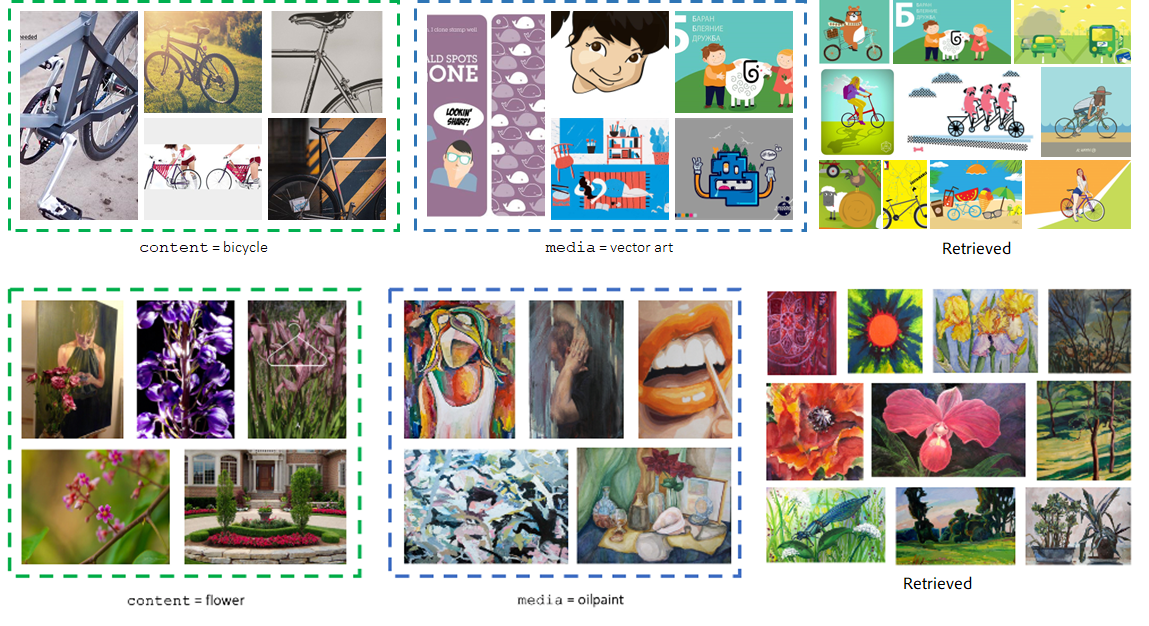}
\caption{\edit{Qualitative results for composing collections using disentangled multi-view representations. (Top) The left subplot represents a collection with the \textit{object} intent of bicycles and the center subplot represents a collection with the \textit{style} intent of vector art images. Disentangled multi-view representations allows constructing a collection query resulting in a natural combined set of results -- bicycles styled in vector art form. (Bottom) The left subplot is a collection with the \textit{object} intent of flowers, while the center subplot is a collection with the \textit{style} intent of oilpaint images. A query formed by composing the relevant disentangled views from these collections retrieves flowers styled in oilpaint form.}}
\label{fig:collCompose}
\end{figure*}

\section{Composing Collections}\label{sec:composing-collections}
\edit{
Deriving disentangled multi-view representations for a collection of images enables the novel use-case of composing multiple collections as a query to retrieve images that selectively adhere to the collections in the query. By picking desired view representations from each collection in the query, we can create a composite representation for a new (hypothetical) collection which can be used as a query for expansion. Figure~\ref{fig:collCompose} illustrates this idea qualitatively using two examples. Consider the top row from Figure~\ref{fig:collCompose}. The query comprises a pair of collections (shown as outlined boxes) and the view that is relevant for each collection. Specifically, we consider the object view from the collection of bicycles and the style view from the collection of vector art images. By selecting the object representations from the former, style representations from the latter, and averaging out the color representations between the two, we obtain composite representations for a hypothetical collection that has the object features of the former and style features of the latter. Since we are only interested in the style and object views, we can split the intent weights between only these for the composite collection before ranking images from the index. Proceeding as described in Section~\ref{sec:collectionbasedRetrieval}, we obtain the ranked list of images shown on the top-right in Figure~\ref{fig:collCompose}, which are images of bicycles styled in vector art form. The disentanglement process described in our paper is critical to enable this behavior. By ensuring that representations along different views are de-correlated, these views can be mixed and matched, allowing for a powerful visual querying mechanism. A similar example, representing the intent of flower images in oilpaint style, is shown in the bottom row of  Figure~\ref{fig:collCompose}.
}

\section{Conclusion and Discussion}
In this work, we have introduced the notion of multi-view representations for image collections and enumerated a well-known set of image similarity axes as views -- object, style, color. The baseline multi-view representation of an image is taken be to a union of popular feature extractors for each view. Our primary contribution is in transforming these input representations to minimize correlations among the views using a self-supervised approach. We have shown that this leads to output representations that better capture the overall characteristics of an image. 

To illustrate the benefits of our approach, we have defined a novel collection level task involving retrieval of images relevant to a set of seed images. We have defined the intent of a collection of images to be a distribution over views -- a higher weight assigned to a view indicates greater homogeneity for that view across images in the collection. Finally, we have shown that using these intent weights allows us to effectively score candidate images with respect to the query collection.

\edit{We have also proposed a new querying mechanism for image search driven by composing multiple collections of images. This is enabled using the ideas and techniques presented in this paper such as \textit{views} of images and representations and \textit{intents} of collections. While we have presented qualitative results here, future work can investigate this quantitatively on datasets more suited for this task.}

The work described here is related to the active topic of representation learning. We have borrowed intuitions like factorized representations~\cite{tsai2018learning} and disentanglement~\cite{zaidi2020measuring} from the domain of NLP and applied it to the setting of image retrieval. As future work, we will look into training our model in an end-to-end manner customized for the retrieval task. We also intend to further evaluate the benefits of our approach via a thorough user study. From the perspective of the application that we have considered, extending our current setup to multiple iterations~\cite{teo2016adaptive,koch2019may} is a natural next step. While the treatment in the current paper is restricted to specific visual axes or views, the proposed framework is generalizable to a broader range of visual representation spaces. Finally, we intend to generalize the benefits of our approach to other datasets as well.

\newpage
\bibliographystyle{ACM-Reference-Format}
\balance
\bibliography{main}

% \appendix
% \input{arxiv/appendix.tex}

\end{document}